\documentstyle[aclap]{article}

\newcommand{\sep}{\,\mid\,}
\newcommand{\Size}[1]{\mbox{$\mid \! {#1} \! \mid$}}

\newcommand{\cfg}{{\rm CFG}}
\newcommand{\mypda}{{\rm PDA}}

\newcommand{\myterm}{{\it\Sigma}}
\newcommand{\nonterm}{N}
\newcommand{\myvar}{V}
\newcommand{\stack}{Q}
\newcommand{\goto}{\mbox{{\it goto\/}}}
\newcommand{\closure}{\mbox{\it closure\/}}
\newcommand{\pred}{\mbox{{\it pred\/}}}

\newcommand{\qin}{q_{\it in}}
\newcommand{\qfin}{q_{\it fin}}

\newtheorem{alg}{Algorithm}
\newtheorem{Th}{Theorem}

\newtheorem{definition}[Th]{Definition}

\newcommand{\pda}[1]{\stackrel{#1}{\mapsto}}

\newcommand{\order}[1]{\mbox{${\cal O}(#1)$}}

\newcommand{\tree}{{\it tree\/}}
\newcommand{\glue}{{\it glue\/}}

\newcommand{\de}{\rightarrow}

\newcommand{\ep}{\epsilon}
\newcommand{\mydot}{\mathrel{\tiny\bullet}}

\newcommand{\cA}{{\cal A}}

\newcommand{\ALR}{{\cal A}_{{\rm LR}}}
\newcommand{\TLR}{T_{{\rm LR}}}

\newcommand{\AtwoLR}{{\cal A_{{\rm 2LR}}}}
\newcommand{\TtwoLR}{T_{{\rm 2LR}}}

\newcommand{\CLR}{\mbox{${\cal C}_{\rm 2LR}$}}

\newcommand{\lrset}{I_{{\rm LR}}}
\newcommand{\TWOlrset}{I_{{\rm 2LR}}}
\newcommand{\lrstate}{{\cal R}_{{\rm LR}}}
\newcommand{\TWOlrstate}{{\cal R}_{{\rm 2LR}}}
\newcommand{\lrsym}{Q_{{\rm LR}}}
\newcommand{\TWOlrsym}{Q_{{\rm 2LR}}}
\newcommand{\TWOrules}{P_{{\rm 2LR}}}

\newcommand{\bos}{\rhd}
\newcommand{\eos}{\lhd}

\newtheorem{ex1}{Example}

\title{\vspace{-0.5in}Efficient Tabular LR Parsing}
\author{
       Mark-Jan Nederhof \\
		Faculty of Arts \\
		University of Groningen \\
		P.O. Box 716 \\
		9700 AS Groningen \\
		The Netherlands \\
                {\tt markjan@let.rug.nl} \And
       Giorgio Satta \\
        	Dipartimento di Elettronica ed Informatica \\ 
        	Universit\`{a} di Padova \\
        	via Gradenigo, 6/A \\
        	I-35131 Padova \\
		Italy \\
                {\tt satta@dei.unipd.it} 
}

\begin{document}
\maketitle

\begin{abstract}
We give a new treatment of tabular LR parsing,
which is an alternative to Tomita's generalized LR algorithm.
The advantage is twofold. Firstly, our treatment is conceptually more
attractive because it uses simpler concepts, such as grammar
transformations and standard tabulation techniques also
know as {\em chart parsing}. Secondly, the static and dynamic
complexity of parsing, both in space and time, is significantly reduced.
\end{abstract}

\section{Introduction}

The efficiency of LR({\it k\/}) parsing techniques \cite{SI90} 
is very attractive from the perspective of natural language 
processing applications.  
This has stimulated the computational linguistics community to 
develop extensions of these techniques to general context-free 
grammar parsing. 
The best-known example is 
{\em generalized\/} LR parsing, also known as
Tomita's algorithm, described by \newcite{TO86}
and further investigated by, for example,
\newcite{TO91a} and \newcite{NE94b}. 
Despite appearances, the graph-structured stacks used to describe 
Tomita's algorithm differ very little from {\em parse tables}, 
or in other words,
generalized LR parsing is one of the
so called
{\em tabular\/} parsing algorithms, among which also the 
CYK algorithm \cite{HA78}
and Earley's algorithm \cite{EA70} can be found. (Tabular parsing is also
known as {\em chart parsing\/}.)

In this paper we investigate the extension of LR parsing to general 
context-free grammars from a more general viewpoint: tabular
algorithms can often be described by the composition
of two constructions. One example is given by \newcite{LA74}
and \newcite{BI89}:
the construction of pushdown automata from grammars and the
simulation of these automata by means of tabulation yield different
tabular algorithms for different such constructions.
Another example, on which our presentation is based, was first
suggested by \newcite{LE89}: a grammar is first transformed
and then a standard tabular algorithm 
along with some filtering condition 
is applied
using the transformed grammar. In our case, the transformation
and the subsequent application of the tabular algorithm result in a new form
of tabular LR parsing.

Our method is more efficient than Tomita's algorithm in two respects.
First, reduce operations are implemented in an efficient way,
by splitting them into several, more primitive, operations
(a similar idea has been proposed by \newcite{KI91} for Tomita's algorithm). 
Second, several paths in the computation that must be 
simulated separately by Tomita's algorithm are collapsed into a single 
computation path, using state minimization techniques. 
Experiments on practical grammars have indicated that there is a 
significant gain in efficiency, with regard to both space and time requirements.

Our grammar transformation produces a so called {\em cover\/} for 
the input 
grammar, which together with the filtering condition 
fully captures the specification of the method, 
abstracting away from algorithmic details such as data structures 
and control flow. 
Since this cover can be easily precomputed,  
implementing our LR parser simply amounts to
running the standard tabular algorithm.
This is very attractive from an application-oriented perspective,
since many actual systems for natural language processing are 
based on these kinds of parsing algorithm.

The remainder of this paper is organized as follows.
In Section~\ref{s:defs} some preliminaries are discussed.
We review the notion of LR automaton in Section~\ref{s:lr}
and introduce the notion of 2LR automaton in Section~\ref{s:tlr}.
Then we specify our tabular LR method in Section~\ref{s:tablr},
and provide an analysis of the algorithm in Section~\ref{s:props}.
Finally, some empirical results are given in Section~\ref{s:empiric},
and further discussion of our method is provided in Section~\ref{s:disc}.

\section{Definitions}
\label{s:defs}

Throughout this paper we use standard formal language notation.
We assume that the reader is familiar with context-free
grammar parsing theory \cite{HA78}.  

A context-free grammar ($\cfg$) is a $4$-tuple 
$G = (\myterm,\nonterm,P,S)$, where $\myterm$ and $\nonterm$
are two finite disjoint sets of terminal 
and nonterminal symbols, respectively, $S \in \nonterm$ is the start symbol, 
and $P$ is a finite set of rules. Each rule has the form
$A \de \alpha$ with $A \in \nonterm$ and $\alpha \in \myvar^*$,
where $\myvar$ denotes $\nonterm \cup \myterm$.
The size of $G$, written $\Size{G}$, is defined
as $\sum_{(A \de \alpha)\in P} \Size{A\alpha}$; by \Size{\alpha} we
mean the length of a string of symbols~$\alpha$.

We generally use symbols $A, B, C, \ldots$ to range over $\nonterm\!$,
symbols $a,b,c, \ldots$ to range over $\myterm\!$,
symbols $X, Y, Z$ to range over $\myvar\!$,
symbols $\alpha, \beta, \gamma, \ldots$ to range over $\myvar^*\!$,
and symbols $v, w, x, \ldots$ to range over $\myterm^*\!$.
We write $\ep$ to denote the empty string.

A $\cfg$ is said to be in {\em binary form\/} if 
$\alpha \in \{\epsilon\} \cup \myvar\cup \nonterm^2$
for all of its rules
$A\de\alpha$. (The binary form does not limit
the (weak) generative capacity of context-free 
grammars \cite{HA78}.)
For technical reasons, we sometimes use the {\em augmented\/} grammar 
associated with $G$, defined as 
$G^\dagger = (\myterm^\dagger,\nonterm^\dagger,P^\dagger,S^\dagger)$,
where $S^\dagger$, $\bos$ and $\eos$ are fresh symbols, 
$\myterm^\dagger=\myterm\cup\{\bos,\eos\}$,
$\nonterm^\dagger=\nonterm\cup\{S^\dagger\}$ and 
$P^\dagger=P\cup\{S^\dagger\de \bos S\eos\}$.

A pushdown automaton ($\mypda$)  is a $5$-tuple
$\cA = (\myterm, \stack, T, \qin, \qfin)$, where $\myterm$,
$\stack$ and $T$ are finite sets of input symbols, stack symbols
and transitions, respectively;
$\qin\in \stack$ is the initial stack symbol and
$\qfin\in \stack$ is the final stack symbol.%
\footnote{
We dispense with the notion of {\em state}, 
traditionally incorporated in the definition of $\mypda$.
This does not affect the power of these devices, since
states can be encoded within stack symbols and transitions.}
Each transition has the form $\delta_1\pda{z} \delta_2$, 
where $\delta_1, \delta_2 \in \stack^*$, 
$1\leq\Size{\delta_1}$, $1\leq\Size{\delta_2}\leq 2$, 
and $z=\ep$ or $z=a$.
We generally use symbols $q, r, s, \ldots$ to range over $\stack$, and
the symbol $\delta$ to range over $\stack^*$. 

Consider a fixed input string $v \in \myterm^*$.
A {\em configuration\/} of the automaton is a pair $(\delta,w)$ consisting
of a stack $\delta \in \stack^*$ and the remaining input $w$,
which is a suffix of the input string $v$.
The rightmost symbol of $\delta$ represents the top of the stack.
The {\em initial\/} configuration has the form
$(\qin, v )$, where the stack is formed by the initial
stack symbol.  
The {\em final\/} configuration has the form
$(\qin\;\qfin, \ep)$, where the stack is formed by the final
stack symbol stacked upon the initial stack symbol. 

The application of a transition $\delta_1\pda{z} \delta_2$
is described as follows.
If the top-most symbols of the stack are $\delta_1$, then these 
symbols may be replaced by $\delta_2$, 
provided that either $z=\ep$, or $z=a$ and $a$ is
the first symbol of the remaining input. Furthermore, if $z=a$ then $a$ is
removed from the remaining input. 
Formally, for a fixed $\mypda$ $\cA$ we define the binary relation 
$\vdash$ on configurations as the least relation satisfying
$(\delta \delta_1,w)\vdash(\delta \delta_2,w)$ if there is a transition
$ \delta_1 \pda{\ep} \delta_2$, and
$(\delta \delta_1,aw)\vdash(\delta \delta_2,w)$ if there is a transition
$ \delta_1 \pda{a} \delta_2$.
The recognition of a certain input $v$ is obtained if starting from the
initial configuration for that input we can reach the final
configuration by repeated application of transitions, or,
formally, if $(\qin,v) \vdash^* (\qin \;\qfin,\epsilon)$,
where $\vdash^*$ denotes the reflexive and transitive closure of
$\vdash$.

By a {\em computation\/} of a PDA we mean a sequence
$(\qin,v)$ $\vdash$ $(\delta_1,w_1)$ $\vdash$ \ldots\ $\vdash$ 
$(\delta_n,w_n)$, $n \geq 0$.
A $\mypda$ is called {\em deterministic\/} 
if for all possible configurations at most one transition is applicable. 
A $\mypda$ is said to be in {\em binary form\/}
if, for all transitions $\delta_1 \pda{z} \delta_2$, 
we have $\Size{\delta_1} \leq 2$.

\section{LR automata}
\label{s:lr}

Let $G = (\myterm, \nonterm, P, S)$ be a $\cfg$.
We recall the notion of LR automaton, which is
a particular kind of $\mypda$.
We make use of the augmented grammar 
$G^\dagger = (\myterm^\dagger, \nonterm^\dagger, P^\dagger, S^\dagger)$ 
introduced in Section~\ref{s:defs}.

Let $\lrset = \{A \de \alpha \mydot \beta \sep (A \de \alpha \beta)
\in P^\dagger \}$.
We introduce the function $\closure$ from $2^{\lrset}$ to 
$2^{\lrset}$ and the function $\goto$ from 
$2^{\lrset} \times \myvar$ to $2^{\lrset}$. 
For any $q \subseteq \lrset$, 
$\closure(q)$ is the smallest set such that
\begin{enumerate}
\item $q \subseteq \closure(q)$; and
\item $(B \de \alpha \mydot A \beta) \in \closure(q)$ and 
$(A\de\gamma)\in P^\dagger$ 
together imply $(A \de\ \mydot \gamma) \in \closure(q)$.
\end{enumerate}
We then define 
\begin{eqnarray*}
\lefteqn{\goto(q,X) =} \\
& & \{ A \de \alpha X \mydot \beta \sep  
		(A \de \alpha \mydot X \beta) \in \closure(q) \}. 
\label{l:goto}
\end{eqnarray*}
We construct a finite set $\lrstate$ as the smallest collection of sets 
satisfying the conditions:
\begin{enumerate}
\item	$\{ S^\dagger \de \bos \mydot S\eos \} \in \lrstate$; and
\item	for every 
	$q \in \lrstate$ and $X \in \myvar$, we have
        $\goto(q,X) \in \lrstate$, provided $\goto(q,X) \neq\emptyset$.
\end{enumerate}
Two elements from $\lrstate$ deserve special attention:
$\qin = \{ S^\dagger \de\bos \mydot S\eos \}$, and $\qfin$, which is defined to be
the unique set in $\lrstate$ containing $(S^\dagger \de \bos S\mydot\eos)$;
in other words, $\qfin=\goto(\qin,S)$.

For $A \in \nonterm$, an $A$-{\em redex\/} is a string 
$q_0 q_1 q_2 \cdots q_m$, $m \geq 0$, of elements from $\lrstate$,
satisfying the following conditions: 
\begin{enumerate}
\item 	$(A \de \alpha \mydot) \in \closure(q_m)$, for some 
	$\alpha = X_1 X_2 \cdots X_m$; and
\item	$\goto(q_{k-1},X_{k}) = q_{k}$, 
	for $1 \leq k \leq m$. 
\end{enumerate}
Note that in such an $A$-redex, $(A \de\ \mydot X_1 X_2 \cdots X_m)$ $\in$
$\closure(q_0)$, and $(A \de X_1\cdots X_k \mydot X_{k+1} \cdots X_m)$ $\in$
$q_k$, for $0<k\leq m$.

The LR automaton associated with $G$ is now introduced.
\begin{definition}
\label{PDA:LR}
$\ALR = (\myterm,\lrsym,\TLR,\qin,\qfin)$, 
where $\lrsym = \lrstate$, $\qin = \{ S^\dagger \de\bos \mydot S\eos \}$, 
$\qfin=\goto(\qin,S)$, and $\TLR$ contains:
\begin{enumerate}
\item
$q \pda{a} {q\; q'} $,
for every $a\in \myterm$ and $q,q'\in \lrstate$ such that $q' = \goto(q, a)$;
\item
${q\delta} \pda{\epsilon} {q\; q'}$,
for every $A\in \nonterm$, $A$-redex $q\delta$, and $q'\in \lrstate$ such that
$q' = \goto(q, A)$.
\end{enumerate}
\end{definition}
Transitions in~(i) above are called {\em shift}, 
transitions in~(ii) are called {\em reduce}.

\section{2LR Automata}
\label{s:tlr}

The automata $\ALR$ defined in the previous section
are deterministic only for a subset of the $\cfg$s, 
called the LR$(0)$ grammars \cite{SI90},
and behave nondeterministically in the general case.
When designing tabular methods that 
simulate nondeterministic computations of $\ALR$, 
two main difficulties are encountered: 
\begin{itemize}
\item A reduce transition in $\ALR$ is an elementary operation 
that removes from the stack a number of elements bounded by 
the size of the underlying grammar.  Consequently, 
the time requirement of tabular
simulation of $\ALR$ computations can be onerous, for 
reasons pointed out by \newcite{SH76} and \newcite{KI91}. 
\item  The set $\lrstate$ can be exponential in the size 
of the grammar \cite{JO91}.  
If in such a case the computations of $\ALR$ touch upon each state, 
then time and space requirements of tabular simulation are 
obviously onerous. 
\end{itemize}
The first issue above is solved here by recasting $\ALR$ 
in binary form. This is done by considering each reduce transition 
as a sequence of ``pop'' operations which affect at most two stack symbols 
at a time. (See also \newcite{LA74}, \newcite{VI93a} and \newcite{NE94b}, 
and for LR parsing specifically \newcite{KI91} and \newcite{LE92a}.)
The following definition introduces this new kind of automaton.

\begin{definition}
$\ALR' = (\myterm, \lrsym', \TLR', \qin, \qfin)$, where
$\lrsym' = \lrstate \cup \lrset$,
$\qin = \{ S^\dagger \de\bos \mydot S\eos \}$, $\qfin=\goto(\qin,S)$
and $\TLR'$ contains:
\begin{enumerate}
\item 
$q \pda{a} {q\; q'} $,
for every $a\in \myterm$ and $q,q'\in \lrstate$ such that $q' = \goto(q, a)$;
\item
$q \pda{\epsilon} {q \;  (A\de \alpha \mydot)}$,
for every $q\in \lrstate$ and $(A\de \alpha \mydot) \in \closure(q)$;
\item
${q \; (A\de \alpha X \mydot \beta)} 
         \pda{\epsilon} (A\de \alpha \mydot X \beta)$,
for every $q\in \lrstate$ and $(A\de \alpha X \mydot \beta) \in q$;
\item
${q \; (A\de\ \mydot\alpha)} \pda{\epsilon} {q\; q'}$,
for every $q,q'\in \lrstate$ and $(A\de\alpha) \in P^\dagger$ such that
$q' = \goto(q, A)$.
\end{enumerate}
\end{definition}

\noindent
Transitions in~(i) above are again called {\em shift}, transitions
in~(ii) are called {\em initiate}, those in~(iii) are called 
{\em gathering}, and transitions in~(iv) are called {\em goto}.
The role of a reduce step in $\ALR$ is taken over in $\ALR'$
by an initiate step, a number of gathering steps, and a goto step.
Observe that these steps involve the new stack symbols
$(A \de \alpha \mydot \beta) \in \lrset$ 
that are distinguishable from possible stack symbols
$\{ A \de \alpha \mydot \beta \} \in \lrstate$. 

We now turn to the second above-mentioned problem, 
regarding the size of set $\lrstate$. 
The problem is in part solved here as follows. 
The number of states in $\lrstate$ 
is considerably reduced by identifying two states if they become identical
after items $A\de\alpha\mydot\beta$ from $\lrset$ have been
simplified to only the suffix of the right-hand side $\beta$.
This is reminiscent of
techniques of state minimization for finite automata \cite{BO67}, 
as they have been applied before to LR parsing, e.g., 
by \newcite{PA70} and \newcite{NE93f}.

Let $G^\dagger$ be the augmented grammar associated with 
a $\cfg$ $G$, and let 
$\TWOlrset = \{\beta \sep (A \de \alpha \beta) \in P^\dagger \}$.
We define variants of the $\closure$ and $\goto$ functions from the
previous section as follows.
For any set $q \subseteq \TWOlrset$, 
$\closure'(q)$ is the smallest collection of sets such that
\begin{enumerate}
\item $q \subseteq \closure'(q)$; and
\item $(A \beta) \in \closure'(q)$ and
$(A\de\gamma) \in P^\dagger$ together imply $(\gamma) \in \closure'(q)$.
\end{enumerate}
Also, we define 
\begin{eqnarray*}
\goto'(q,X) & = & \{ \beta \sep (X\beta) \in \closure'(q) \}.
\label{l:gotoprime}
\end{eqnarray*}
We now construct a finite set $\TWOlrstate$ as the smallest set 
satisfying the conditions:
\begin{enumerate}
\item   $\{ S\eos \} \in \TWOlrstate$; and 
\item   for every
        $q \in \TWOlrstate$ and $X \in \myvar$, we have
        $\goto'(q,X) \in \TWOlrstate$, provided $\goto'(q,X) \neq\emptyset$.
\end{enumerate} 

As stack symbols, we take the elements from $\TWOlrset$ and a subset
of elements from $(\myvar\times\TWOlrstate)$:
\begin{eqnarray*}
\TWOlrsym &\!=\!& \{(X, q)\sep \exists q'[\goto'(q',X)=q]\}\ \cup \TWOlrset
\end{eqnarray*}
In a stack symbol of the form $(X, q)$,
the $X$ serves to record the grammar symbol
that has been recognized last, cf.\ the symbols that formerly were found
immediately before the dots. 

The 2LR automaton associated with $G$ can now be introduced. 
\begin{definition}
\label{d:rtwolr}
$\AtwoLR = (\myterm, \TWOlrsym, \TtwoLR, \qin', \qfin')$, where
$\TWOlrsym$ is as defined above,
$\qin'=(\bos,\{S\eos\})$, $\qfin'=(S,\goto'(\{S\eos\},S))$, 
and $\TtwoLR$ contains:
\begin{enumerate}
\item $(X,q) \pda{a} {(X,q)\; (a,q')}$,
for every $a \in \myterm$ and $(X,q),(a,q') \in \TWOlrsym$ such that
$q' = \goto'(q, a)$;
\item 
$(X,q) \pda{\epsilon} {(X, q) \; (\ep)}$,
for every $(X, q) \in \TWOlrsym$ such that $\ep\in \closure'(q)$; 
\item 
${(X,q) \; (\beta)} \pda{\epsilon} (X \beta)$,
for every $(X,q)\in \TWOlrsym$ and $\beta \in q$; 
\item
${(X,q) \; (\alpha)} \pda{\epsilon} {(X,q)\; (A,q')}$, 
for every $(X,q)$, $(A,q') \in \TWOlrsym$ and $(A \de \alpha) \in P^\dagger$
such that $q' = \goto'(q, A)$.
\end{enumerate}
\end{definition}
Note that in the case of a reduce/reduce conflict with two grammar
rules sharing some suffix in the right-hand side, the gathering steps
of $\AtwoLR$ will treat both rules simultaneously,
until the parts of the right-hand sides are reached where the two rules differ.
(See \newcite{LE92} for a similar sharing of computation for
common suffixes.)

An interesting fact is that the automaton $\AtwoLR$ is very similar
to the automaton $\ALR$ constructed for a grammar transformed
by the transformation $\tau_{\it two}$ given by \newcite{NE94}.%
\footnote{For the earliest mention of this transformation, we have
encountered pointers to \newcite{SC73a}. Regrettably, we have as yet not been
able to get hold of a copy of this paper.}

\section{The algorithm}
\label{s:tablr}

This section presents a tabular LR parser, 
which is the main result of this paper. 
The parser is derived from the 2LR automata 
introduced in the previous section.  Following the general approach 
presented by \newcite{LE89}, we simulate computations of these devices 
using a tabular method, a grammar transformation and a filtering
function.

We make use of a tabular parsing algorithm which is basically 
an asynchronous version of the CYK algorithm, 
as presented by \newcite{HA78}, 
extended to productions of the forms $A \de B$ and $A\de\epsilon$ and 
with a left-to-right filtering condition. 
The algorithm uses a parse table consisting in a $0$-indexed 
square array $U$. The indices represent positions in the input string.
We define $U_i$ to be $\bigcup_{k \leq i} U_{k,i}$.

Computation of the entries of $U$ is moderated by a filtering process.
This process makes use of a function $\pred$ from
$2^{\nonterm}$ to $2^{\nonterm}$, specific to a certain
context-free grammar.
We have a certain nonterminal $A_{\it init}$ which is initially
inserted in $U_{0,0}$ in order to start the recognition process.

We are now ready to give a formal specification of the tabular algorithm.

\begin{alg} \rm
\label{a:cyk}
Let $G = (\myterm, \nonterm, P, S)$ be a $\cfg$ in binary form,
let $\pred$ be a function from
$2^{\nonterm}$ to $2^{\nonterm}$, let $A_{\it init}$ be the distinguished 
element from $\nonterm$,
and let $v = a_1 a_2 \cdots a_n \in \myterm^*$ be an input string.
We compute the least $(n+1) \times (n+1)$ 
table $U$ such that $A_{\it init} \in U_{0,0}$ and 
\begin{enumerate}
\item $A \in U_{j-1,j}$ \\
      \ \ \ \ if $(A \de a_j)\in P$, $A \in \pred(U_{j-1})$; 
\item $A \in U_{j,j}$ \\
      \ \ \ \ if $(A \de \epsilon)\in P$, $A \in \pred(U_{j})$; 
\item $A \in U_{i,j}$ \\
      \ \ \ \ if $B \in U_{i,k}$, $C \in U_{k,j}$, $(A \de BC)\in P$, 
	$A \in \pred(U_i)$;
\item $A \in U_{i,j}$ \\
      \ \ \ \ if $B \in U_{i,j}$, $(A \de B)\in P$, $A \in \pred(U_i)$.
\end{enumerate}
\end{alg}
The string has been accepted when $S \in U_{0,n}$.

We now specify a grammar transformation, based on the 
definition of $\AtwoLR$.  

\begin{definition}
\label{d:lrcover}
Let $\AtwoLR = (\myterm, \TWOlrsym, \TtwoLR, \qin', \qfin')$
be the $2${\rm LR} automaton associated with 
a $\cfg$ $G$. 
The {\bf 2LR cover\/} associated with $G$ is the $\cfg$ 
$\CLR(G) = (\myterm, \TWOlrsym, \TWOrules, \qfin')$,
where the rules in $\TWOrules$ are given by:
\begin{enumerate}
\item   $(a,q') \de a$,\\
        for every $(X,q) \pda{a} {(X,q)\; (a,q')} \in \TtwoLR$; 
\item   $(\epsilon) \de \epsilon$,\\
        for every $(X,q) \pda{\epsilon} (X,q) \; (\epsilon) \in \TtwoLR$; 
\item   $(X \beta) \de (X,q)\;(\beta)$,\\
        for every 
        ${(X,q)\;(\beta)} \pda{\epsilon} (X \beta) \in \TtwoLR$; 
\item   $(A,q') \de (\alpha)$,\\
    for every ${(X,q)\;(\alpha)} \pda{\epsilon} {(X,q)\; (A,q')} \in \TtwoLR$. 
\end{enumerate}
\end{definition}
Observe that there is a direct, one-to-one correspondence between 
transitions of $\AtwoLR$ and productions of $\CLR(G)$.  

The accompanying function $\pred$ is defined as follows
($q,q',q''$ range over the stack elements):
\begin{eqnarray*}
\pred(\tau) & = & 
\{ q\sep {q' q''} \pda{\epsilon} q \in \TtwoLR\}\
       \cup \\
&& \{ q\sep q' \in\tau,\; q' \pda{z} {q'\; q} \in \TtwoLR\}\
       \cup \\
&& \{ q\sep q'\in\tau,\; {q'\;q''} \pda{\epsilon} {q'\; q}
                                       \in \TtwoLR\}. 
\end{eqnarray*} 
The above definition implies that only the tabular equivalents of the
shift, initiate and goto transitions are subject to actual filtering; 
the simulation
of the gathering transitions does not depend on elements in $\tau$.

Finally, the distinguished nonterminal from the cover used to initialize 
the table is $\qin'$. Thus we start with $(\bos,\{S\eos\}) \in U_{0,0}$.

The 2LR cover introduces spurious ambiguity: where some grammar
$G$ would allow a certain number of parses to be found for a certain
input, the grammar $\CLR(G)$ in general allows {\em more\/} parses.
This problem is in part solved by the filtering function $\pred$.
The remaining spurious ambiguity is avoided by a particular
way of constructing the parse trees, described in what follows.

After Algorithm~\ref{a:cyk} has recognized a given input,
the set of all parse trees can be computed as 
$\tree(\qfin',0,n)$ where the function $\tree$, which 
determines sets of either parse trees or lists of parse trees
for entries in $U$, is recursively defined by:
\begin{enumerate}
\item $\tree((a,q'),i,j)$ is the set $\{a\}$. This set contains a single
parse tree consisting of a single node labelled $a$.
\item $\tree(\epsilon,i,i)$ is the set $\{\epsilon\}$. This set consists of
an empty list of trees.
\item $\tree(X\beta,i,j)$ is the union of the sets 
${\cal T}_{(X\beta),i,j}^k$,
where $i\leq k\leq j$, $(\beta) \in U_{k,j}$, 
and there is at least one
$(X,q) \in U_{i,k}$ and
$(X \beta) \de (X,q)\;(\beta)$ in $\CLR(G)$, for some $q$.
For each such $k$, select one such $q$. We define
${\cal T}_{(X\beta),i,j}^k = 
\{t \cdot {\it ts} \sep t \in \tree((X,q),i,k) \wedge
{\it ts} \in \tree(\beta,k,j)\}$. Each $t \cdot {\it ts}$ is a list of
trees, with head $t$ and tail ${\it ts}$.
\item $\tree((A,q'),i,j)$ is the union of the sets 
${\cal T}_{(A,q'),i,j}^{\alpha}$,
where $(\alpha) \in U_{i,j}$ is such that 
$(A,q') \de (\alpha)$ in $\CLR(G)$.
We define ${\cal T}_{(A,q'),i,j}^{\alpha}=\{glue(A,{\it ts}) \sep 
{\it ts}\in \tree(\alpha,i,j)\}$.
The function $\glue$ constructs a tree from a fresh root node labelled
$A$ and the trees in list ${\it ts}$ as immediate subtrees.
\end{enumerate}
We emphasize that in the third clause above, 
one should not consider more than one 
$q$ for given $k$ in order to prevent spurious ambiguity.
(In fact, for fixed $X,i,k$ and for different $q$ 
such that $(X,q) \in U_{i,k}$, $\tree((X,q),i,k)$ yields 
the exact same set of trees.)
With this proviso, the degree of ambiguity, i.e.\ the number of parses
found by the algorithm for any input, is reduced to exactly that of 
the source grammar.

A practical implementation would construct the parse trees on-the-fly,
attaching them to the table entries, allowing packing and sharing
of subtrees (cf.\ the literature on {\em parse forests\/} 
\cite{TO86,BI89}). 
Our algorithm 
actually only needs one (packed) subtree for several $(X,q)\in U_{i,k}$ with 
fixed $X,i,k$ but different $q$.
The resulting parse forests would then be optimally
compact, contrary to some other LR-based tabular algorithms, as pointed
out by \newcite{RE92}, \newcite{NE93b} and \newcite{NE94a}. 

\section{Analysis of the algorithm}
\label{s:props}

In this section, we investigate how the steps performed 
by Algorithm~\ref{a:cyk} (applied to the 2LR cover) relate
to those performed by $\AtwoLR$, for the same input.

We define a subrelation $\models^+$ of $\vdash^+$ as:
$(\delta,uw)\models^+(\delta\delta',w)$ if and only if
$(\delta,uw)=(\delta,z_1z_2\cdots z_mw)
\vdash(\delta\delta_1,z_2\cdots z_mw)\vdash\ldots\vdash(\delta\delta_m,w)=
(\delta\delta',w)$,
for some $m\geq 1$, where $|\delta_k|>0$ for all $k$, $1\leq k \leq m$.
Informally, we have $(\delta,uw)\models^+(\delta\delta',w)$ if configuration
$(\delta\delta',w)$ can be reached from $(\delta,uw)$ without the
bottom-most part $\delta$ of the intermediate stacks being affected by
any of the transitions;
furthermore, at least one element is pushed on top of $\delta$.

The following characterization relates  
the automaton $\AtwoLR$ and Algorithm~\ref{a:cyk} applied to the
2LR cover.
Symbol $q\in\TWOlrsym$ is eventually 
added to $U_{i,j}$ if and only if for some $\delta$:
$$(\qin',a_1\ldots a_n) \vdash^* (\delta,a_{i+1}\ldots a_n)
\models^+ (\delta q,a_{j+1}\ldots a_n).$$ 
In words, $q$ is found in entry $U_{i,j}$ if and only if, 
at input position $j$, the automaton would push some 
element $q$ on top of some lower-part of the stack $\delta$ that 
remains unaffected while the input from $i$ to $j$ is being read.

The above characterization, whose proof is not reported here,
is the justification for calling the resulting algorithm 
tabular LR parsing. In particular, for
a grammar for which $\AtwoLR$ is deterministic, i.e.\ for an
LR(0) grammar, the number of steps performed by $\AtwoLR$ and
the number of steps performed by the above algorithm 
are exactly the same. In the case
of grammars which are not LR(0), the tabular LR algorithm
is more efficient than for example a backtrack realisation of $\AtwoLR$.

For determining the order of the time complexity of
our algorithm, we look at the most
expensive step, which
is the computation of an element $(X\beta) \in U_{i,j}$ from 
two elements $(X,q) \in U_{i,k}$ and $(\beta) \in U_{k,j}$, through
${(X,q)\;(\beta)} \pda{\epsilon} (X \beta) \in \TtwoLR$.
In a straightforward realisation of the algorithm,
this step can be applied $\order{\Size{\TtwoLR}\cdot\Size{v}^3}$ 
times (once for each $i,k,j$ and each transition), each step taking
a constant amount of time.
We conclude that the time complexity of our algorithm
is $\order{\Size{\TtwoLR}\cdot\Size{v}^3}$. 

As far as space requirements are concerned, each set $U_{i,j}$
or $U_i$ contains at most $\Size{\TWOlrsym}$ elements.
(One may assume an auxiliary table storing each $U_i$.)
This results in a space complexity 
$\order{\Size{\TWOlrsym}\cdot\Size{v}^2}$.

The entries in the table represent single stack elements, as opposed to
pairs of stack elements following \newcite{LA74} and \newcite{LE89}. 
This has been investigated before 
by \newcite[p.\ 25] {NE94b} and \newcite[p.\ 155]{VI93a}.

\section{Empirical results}
\label{s:empiric}

We have performed some experiments with Algorithm~\ref{a:cyk} applied to
$\AtwoLR$ and $\ALR'$, for 4 practical context-free grammars.
For $\ALR'$ a cover was used analogous to 
the one in 
Definition~\ref{d:lrcover};
the filtering function remains the same.

The first grammar 
generates a subset of the programming language ALGOL~68
\cite{vW75}. The second and third 
grammars generate a fragment of Dutch, and are 
referred to 
as the CORRie grammar \cite{VO94} and the Deltra
grammar \cite{SH90}, 
respectively. 
These grammars were stripped of their arguments
in order to 
convert them into context-free grammars.  
The fourth grammar, referred to as the
Alvey grammar \cite{CA93}, 
generates a fragment of English and was
automatically generated from a unification-based grammar.

The test sentences
have been obtained by automatic generation from the grammars,
using the Grammar Workbench \cite{NE92c}, which uses a random
generator to select rules; 
therefore these sentences do 
not necessarily represent input typical 
of the applications 
for which the grammars were written. 
Table~\ref{grammars} summarizes the test material. 
\begin{table}[t]
\begin{center}
\renewcommand{\arraystretch}{1.2}
\begin{tabular}{|l||r|r|r||r|}
\hline
$G = (\myterm,\nonterm,P,S)$ & 
\multicolumn{1}{c|}{$\Size{G}$} &
\multicolumn{1}{c|}{$|\nonterm|$}&
\multicolumn{1}{c|}{$|P|$} & 
\multicolumn{1}{c|}{$|w|$} \\
\hline
\hline
ALGOL 68 & 783 & 167 & 330 & 13.7  \\
CORRie   & 1141 & 203 & 424 & 12.3  \\
Deltra   & 1929 & 281 & 703 & 10.8  \\
Alvey    & 5072 & 265 & 1484 & 10.7   \\
\hline
\end{tabular}
\end{center}
\caption{The test material: the four grammars and some of their
dimensions, and the average length of 
the test sentences (20 sentences of various length for each grammar).}
\label{grammars}
\end{table}

Our implementation is merely a prototype, which means that
absolute duration of the parsing process is little indicative of
the actual efficiency of more sophisticated implementations.
Therefore, our measurements have been restricted to implementation-independent
quantities, viz.\ the number of elements stored
in the parse table and the number of elementary steps
performed by the algorithm. In a
practical implementation, such quantities will strongly influence the 
space and time complexity, although they do not represent the only
determining factors. Furthermore, all optimizations
of the time and space efficiency have been left out of consideration.

Table~\ref{results} presents the costs of parsing the 
test sentences.
The first and third columns 
give the number of entries stored in table
$U$, the second 
and fourth 
columns give the number of elementary steps that were
performed. 

An elementary step consists of the derivation of one element
in $\lrsym'$ or $\TWOlrsym$ from one or two 
other elements. 
The elements that are used in the filtering process are counted
individually. 
We give an example for the case of $\ALR'$.
Suppose we derive an element $q'\in U_{i,j}$ from
an element $(A\de\ \mydot\alpha) \in U_{i,j}$, warranted by
two elements $q_1,q_2\in U_i$, $q_1\neq q_2$,
through $\pred$, in the presence of
${q_1 \; (A\de\ \mydot\alpha)} \pda{\epsilon} {q_1\; q'} \in\TLR'$ and
${q_2 \; (A\de\ \mydot\alpha)} \pda{\epsilon} {q_2\; q'} \in\TLR'$.
We then count 
{\em two\/} parsing steps, one for $q_1$ and one for $q_2$.

\begin{table}[t]
\begin{center}
\renewcommand{\arraystretch}{1.2}
\begin{tabular}{|l||r|r||r|r|}
\hline
    & \multicolumn{2}{c||}{$\ALR'$}  & \multicolumn{2}{|c|}{$\AtwoLR$} \\
\cline{2-5} 
$G$ &  \multicolumn{1}{c|}{space} & 
       \multicolumn{1}{c||}{time} &
       \multicolumn{1}{c|}{space} & 
       \multicolumn{1}{c|}{time}\\
\hline
\hline
ALGOL 68 & 327 & 375 & 234 & 343 \\
CORRie & 7548 & 28028 & 5131 & 22414  \\
Deltra   & 11772 & 94824 & 6526 & 70333 \\
Alvey   & 599 & 1147 & 354 & 747 \\
\hline
\end{tabular}
\end{center}
\caption{Dynamic requirements: average space and time per sentence.}
\vspace{-.5em}
\label{results}
\end{table}
\begin{table*}[tb]
\begin{center}
\renewcommand{\arraystretch}{1.2}
\begin{tabular}{|l||r|r|r||r|r|r|}
\hline 
    & \multicolumn{3}{c||}{$\ALR'$}  & \multicolumn{3}{|c|}{$\AtwoLR$} \\
\cline{2-7}
$G$ & \multicolumn{1}{c|}{$|\lrstate|$} & 
      \multicolumn{1}{c|}{$|\lrsym'|$} &  
      \multicolumn{1}{c||}{$|\TLR'|$} &
      \multicolumn{1}{c|}{$|\TWOlrstate|$} & 
      \multicolumn{1}{c|}{$|\TWOlrsym|$} &  
      \multicolumn{1}{c|}{$|\TtwoLR|$}   \\
\hline
\hline
ALGOL 68 & 434 & 1,217 & 13,844 & 109 & 724 & 12,387 \\
CORRie   & 600 & 1,741 & 22,129 & 185 & 821 & 15,569 \\
Deltra   & 856 & 2,785 & 54,932 & 260 & 1,089 & 37,510 \\
Alvey    & 3,712 & 8,784 & 1,862,492 & 753 & 3,065 & 537,852 \\
\hline
\end{tabular}
\end{center}
\caption{Static requirements.}
\vspace{-.5em}
\label{sizes}
\end{table*}

Table~\ref{results} shows that there is a significant gain in
space and time efficiency 
when moving 
from $\ALR'$ to $\AtwoLR$.

Apart from the dynamic costs of parsing, we have also measured some
quantities relevant to the construction and storage of the two types
of tabular LR parser. 
These data are given in Table~\ref{sizes}.

We see that the number of states is strongly reduced
with regard to traditional LR parsing. In the case of the Alvey
grammar, 
moving from $|\lrstate|$ to $|\TWOlrstate|$ amounts
to a reduction to 20.3~\%. 
Whereas time- and space-efficient computation 
of $\lrstate$ for this grammar is a serious problem,
computation of $\TWOlrstate$ will not be difficult on any modern
computer. Also significant is the reduction from $|\TLR'|$ to
$|\TtwoLR|$, especially for the larger grammars. These quantities 
correlate with the amount of storage needed for naive representation
of the respective automata.

\section{Discussion}
\label{s:disc}

Our treatment of tabular LR parsing has two important advantages over
the one by Tomita: 
\begin{itemize}
\item It is conceptually simpler, because we make use of simple
concepts such as a grammar transformation and the well-understood
CYK algorithm, instead of a complicated mechanism working on 
graph-structured stacks.
\item Our algorithm requires fewer LR states.
This leads to faster parser generation, to smaller parsers, 
and to reduced time and space complexity of parsing itself.
\end{itemize}

The conceptual simplicity of our formulation of tabular
LR parsing allows comparison with other tabular parsing techniques,
such as
Earley's algorithm \cite{EA70} and tabular left-corner
parsing \cite{NE93b}, based on implementation-independent criteria.
This is in contrast to experiments reported before (e.g.\ by
\newcite{SH91}),
which treated tabular LR parsing differently from the other
techniques.

The reduced time and space complexities reported in the previous
section pertain to the tabular realisation of two
parsing techniques, expressed by the automata $\ALR'$ and 
$\AtwoLR$. The tabular realisation of the former automata is
very close to a variant of Tomita's algorithm 
by \newcite{KI91}. 
The objective of our experiments was to show that the
automata $\AtwoLR$ provide a better basis than $\ALR'$ for 
tabular LR parsing with regard to space and time complexity. 

Parsing algorithms that are not based on the LR technique
have however been left out of consideration, and so
were techniques for unification grammars and techniques incorporating
finite-state processes.%
\footnote{As remarked before by \newcite{NE93b}, the algorithms by
\newcite{SC91} and \newcite{LE89} are not really related to LR parsing, although
some notation used in these papers suggests otherwise.}

Theoretical 
considerations \cite{LE89,SC91,NE94a} have suggested that
for natural language parsing, LR-based techniques may not
necessarily be superior to other parsing techniques, although
convincing empirical data to this effect has never been shown.
This issue is difficult to resolve because so much of the 
relative efficiency of the different parsing techniques depends
on particular grammars and particular input, as well as
on particular implementations of the techniques. We hope
the conceptual framework presented in this paper may
at least partly alleviate this problem.

\section*{Acknowledgements}

The first author is supported by the Dutch Organization
for Scientific Research (NWO), under grant 305-00-802.
Part of the present research was done
while the second author was visiting the
Center for Language and Speech Processing,
Johns Hopkins University, Baltimore, MD.

We received kind help from 
John Carroll, Job Honig, Kees Koster, Theo Vosse and Hans de Vreught
in finding the grammars mentioned in this paper.
Generous help with locating relevant literature was provided by Anton
Nijholt, Rockford Ross, and Arnd Ru{\ss}mann.

\bibliographystyle{fullname}
\newcommand{\noop}[1]{}\newcommand{\id}[1]{#1}

\end{document}